\documentclass[]{article}
\usepackage{amsmath}
\usepackage{amssymb}
\usepackage{graphicx}
\usepackage{placeins}
\usepackage[textwidth=18cm, textheight=24cm]{geometry}
\usepackage{lineno}
\usepackage{dsfont}
\usepackage{hyperref}
\usepackage[round]{natbib}
\usepackage{siunitx}
\usepackage{float}
\usepackage{ccaption}
\usepackage{authblk}
\usepackage{textcomp}
\usepackage{dsfont}
\usepackage{svg}
\usepackage{comment}
\usepackage{bbm}
\newcommand*\samethanks[1][\value{footnote}]{\footnotemark[#1]}

\DeclareMathOperator*{\argmax}{arg\,max}

\title{Neuromorphic Hardware learns to learn}
\author[1]{Thomas Bohnstingl\footnote{equal contribution}\thanks{bohnstingl@student.tugraz.at}}
\author[1]{Franz Scherr\samethanks[1]}
\author[2]{Christian Pehle}
\author[2]{Karlheinz Meier}
\author[1]{Wolfgang Maass}
\affil[1]{Institute for Theoretical Computer Science\\Graz University of Technology, Austria}
\affil[2]{Kirchhoff-Institute for Physics, Ruprecht-Karls-Universit{\"a}t Heidelberg}

\begin{document}

\maketitle

\begin{abstract}
Hyperparameters and learning algorithms for neuromorphic hardware are usually chosen by hand. In contrast, the hyperparameters and learning algorithms of networks of neurons in the brain, which they aim to emulate, have been optimized through extensive evolutionary and developmental processes for specific ranges of computing and learning tasks. Occasionally this process has been emulated through genetic algorithms, but these require themselves hand-design of their details and tend to provide a limited range of improvements. We employ instead other powerful gradient-free optimization tools, such as cross-entropy methods and evolutionary strategies, in order to port the function of biological optimization processes to neuromorphic hardware. As an example, we show that this method produces neuromorphic agents that learn very efficiently from rewards. In particular, meta-plasticity, i.e., the optimization of the learning rule which they use, substantially enhances reward-based learning capability of the hardware. In addition, we demonstrate for the first time Learning-to-Learn benefits from such hardware, in particular, the capability to extract abstract knowledge from prior learning experiences that speeds up the learning of new but related tasks. Learning-to-Learn is especially suited for accelerated neuromorphic hardware, since it makes it feasible to carry out the required very large number of network computations.\\

\end{abstract}

\section{Introduction}
The computational substrate that the human brain employs to carry out its computational functions, is given by networks of spiking neurons (SNNs). There appear to be numerous reasons for evolution to branch off towards such a design. For example, networks of such neurons facilitate a distributed scheme of computation, intertwined with memory entities. Thereby overcoming known disadvantages in contemporary computer designs such as the von Neumann bottleneck. Importantly, the human brain serves as a blueprint for a power efficient learning machine, solving demanding computational tasks while consuming little resources. A characteristic property that makes energy efficient computation possible is the distinct communication among these neurons. In particular, neurons do not need to produce an output at all times. Instead, information is integrated over time and communicated sparsely using a format of discrete events, ``spikes''.\\

\noindent The connectivity structure, the development of computational functions in specific brain regions, as well as the active learning algorithms are all subject to an evolutionary process. In particular, evolution has shaped the human brain and successfully formed a learning machine, capable to carry out a range of complex computations. In close connection to this, a characteristic property of learning processes in humans is the ability to take advantage of previous, related experiences and use them in novel tasks. Indeed, humans show both, the ability to quickly adapt to new challenges in various domains, and the ability to transfer prior acquired knowledge about different, but related tasks to new, potentially unseen ones \citep{Taylor2009, RobertCanini2010, Wang2015}.\\

\noindent One strategy to investigate the benefit of a knowledge transfer between different, but related learning tasks is to impose a so-called Learning-to-Learn (L2L) optimization. L2L employs task-specific learning algorithms, but also tries to mimic the slow evolutionary and developmental processes that have prepared brains for the learning tasks humans have to face. In particular, L2L introduces a nested optimization procedure, consisting of an inner loop and an outer loop. In the inner loop, specific tasks are learned, while an additional outer loop aims to optimize the learning performance on a range of different tasks. This concept gave rise to an interesting body of work \citep{hochreiter, Finn2017, Wang2016} and showed that one can endow artificial learning systems with transfer learning capabilities. Recently, this concept was also extended to networks of spiking neurons. In \citep{Bellec2018} it is shown that a biologically inspired circuit can encode prior assumptions about the tasks it will encounter.\\

\noindent Usually, one takes advantage of the availability of gradient information to facilitate optimization, here instead, we employ powerful gradient-free optimization algorithms in the outer loop that emulate the evolutionary process. In particular, we demonstrate the benefits of evolutionary strategies (ES)~\citep{Rechenberg1973} and cross entropy methods (CE)~\citep{Rubinstein1997}, as they are able to deal with noisy function evaluations and perform in high-dimensional spaces. In the inner loop, on the other hand, we consider reinforcement learning problems (RL problems), such as Markov Decision Processes and Multi-armed bandits. Problems of this type appear quite often in general and therefore, a rich literature has emerged. However, it still remains that learning from rewards is particularly inefficient, as the feedback is given by a single scalar quantity, the reward. We show that by employing the concept of L2L we can produce agents that learn efficiently from rewards and exploit previous experiences on related, new tasks.\\

\noindent As another novelty, we implement the learning agent on a neuromorphic hardware (NM hardware). Specialized hardware of this type has emerged by taking inspiration of principles of brain computation, with the intent to port the advantages of distributed and power efficient computation to silicon chips \citep{mead1990neuromorphic}. This holds the great promise to install artificial intelligence in devices without cloud connection and/or limited resource. Numerous architectures have been proposed that are either based on analog, digital or mixed-signal approaches: \citep{Ambrogio2018, Furber2016, Schemmel2010, Aamir2018, Furber2014, Pantazi2016, Davies2018}. We refer to \citep{Schuman2017} for a survey on neuromorphic systems.\\

\noindent In order to further enhance the learning capabilities of NM hardware, we exploit the adjustability of the employed neuromorphic chip and consider the use of meta-plasticity. In other words, we evolve a highly configurable plasticity rule that is responsible for learning in the network of spiking neurons. To this end, we represent the plasticity rule as a multilayer perceptron and demonstrate that this approach can significantly boost learning performance as compared to the level that is achieved by plasticity rules that we derive from general algorithms.\\

\noindent NM hardware is especially well suited for L2L because it renders the large number of simulations that need to be carried out feasible. Spiking neurons that are simulated on NM hardware typically exhibit accelerated dynamics as compared to their biological counterparts. In addition, the chosen neuromorphic hardware allows to emulate both, the RL environment as well as the learning algorithm at the same acceleration factor and hence, one unlocks the full potential of the specialized neuromorphic chip.\\

\noindent First, in Section~\ref{sec:methods} we will discuss our approaches and methods, as well as the set of tools that was used in our experiments. In particular, the NM hardware that was employed is discussed. Then, in Section~\ref{RLTL} we will exhibit the increase in performance that we obtained on NM hardware for the conducted tasks and discuss which gradient-free algorithms worked best for us. In the following, we discuss in Section~\ref{sec:meta} that performance can be further increased by the adoption of a highly customizable learning rule, i.e. meta-plasticity, that is shaped through L2L, and discuss its relevance in transfer learning. We further discuss the impact in terms of simulation time thanks to the underlying NM hardware. Finally, we conclude our findings and results in Section \ref{sec:discussion}.

\section{Methods and Materials}
\label{sec:methods}
This section provides the technical details to the conducted experiments. First, we describe the background for L2L in Section~\ref{MMLTL}, and discuss the gradient-free optimization techniques that are employed. Subsequently, we provide details to the reinforcement learning tasks that we considered (Section~\ref{MMRL}).\\
\noindent Since the agent that interacts with the RL environments is implemented on a NM hardware, we discuss the corresponding chip in Section~\ref{MMNMHW}. We exhibit the network structure that we used throughout all our experiments in Section~\ref{MMNWStructure}. Subsequently, we provide details to the learning algorithms that we used in Section \ref{MMLA} and discuss methods for analysis.

\subsection{Learning-to-Learn and Gradient-free Optimization}
\label{MMLTL}
The goal of Learning-to-Learn is to enhance a learning systems capability to learn. In models of neural networks, learning performance can be enhanced by several measurements. For example, one can optimize hyperparameters that affect the learning procedure or optimize the learning procedure as such. Often, this optimization is carried out manually and involves a lot of domain knowledge. Here, we approach this problem with L2L and evolve suitable hyperparameters as well as learning algorithms automatically.

\begin{figure}[h]
\centering
\fbox{\includegraphics[width=85mm]{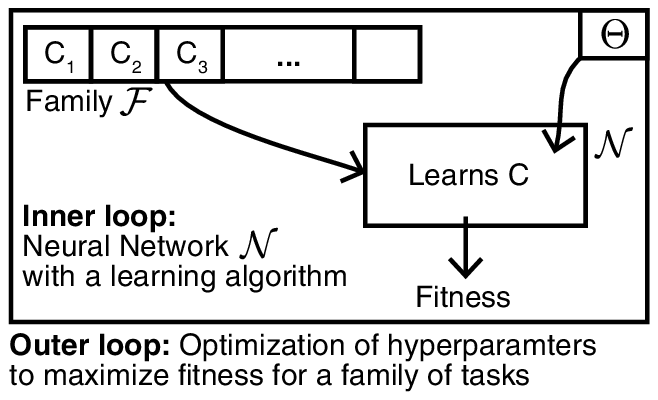}}
\caption{\textbf{Learning-to-Learn scheme}. Learning-to-Learn introduces a nested optimization with two loops. In the inner loop a model is required to perform a learning task $C$ from a family $\mathcal{F}$. The learning capabilities of the model are influenced by hyperparameters $\Theta$ that are optimized in the outer loop in order to maximize a learning fitness across the entire family $\mathcal{F}$.}
\label{fig:methodsLTL}
\end{figure}

\noindent In particular, L2L introduces a nested optimization that consists of two loops: an inner loop and an outer loop as displayed in Figure~\ref{fig:methodsLTL}. In the inner loop, one is solely concerned about learning particular tasks $C$, which are sampled at random from a family of learning tasks $\mathcal{F}$ that share some abstract concepts. The outer loop, on the other hand, is responsible to increase a learning fitness $f$ over many tasks $C$. We express the learning fitness as a function $f(C; \Theta)$ that depends on the task $C$ to be learned, as well as hyperparameters $\Theta$ that characterize the learning algorithm. Formally, we write the goal of L2L as an optimization problem:
\begin{align}
\label{eq:ltl}
    \Theta = \argmax_{\Theta'} \mathds{E}_{C\sim \mathcal{F}} \left[ f(C; \Theta')\right],
\end{align}
and emphasize that $f$ includes the learning process of a task $C$.

\noindent In practice, the expectation in Equation~\eqref{eq:ltl} is approximated using batches of $N$ different tasks: $\mathds{E}_{C\sim \mathcal{F}} \left[ f(C; \Theta) \right] \approx \frac{1}{N} \sum_{i=1}^{N} f(C_i; \Theta)$.
As a result of considering different tasks $C$ in the inner loop each time, the hyperparameters can only assume task independent concepts that are shared throughout the family. In fact, one can consider L2L as an optimization that happens on two different timescales: fast learning of single tasks in the inner loop, and a slower learning process that adapts hyperparameters in order to boost learning on the entire family of learning tasks.\\

\noindent The L2L scheme allows separating the learning process in the inner loop from the optimization algorithms that work in the outer loop. We used Q-Learning and Meta-Plasticity to implement learning in the inner loop (discussed in Section \ref{MMLA}), while at the same time, we considered several gradient-free optimization techniques in the outer loop. The requirements for a well-suited optimization algorithm in the outer loop are the ability to operate in a high-dimensional parameter space, the ability to deal with noisy fitness evaluations, the ability to find a good final solution and the ability to do so using a small number of fitness evaluations. Due to this broad set of requirements, the choice of the outer loop algorithm is non-trivial and needs to be adjusted based on the task family that is considered in the inner loop. We selected a set of gradient-free optimization techniques such as cross-entropy methods, evolutionary strategies, numerical gradient-descent as well as a parallelized variation of simulated annealing. In the following, we provide a brief outline of the algorithms used and refer to the corresponding literature. For the concrete implementation, we employ a L2L software framework that provides several such optimization methods~\citep{subramoney_anand_2019_2553956}. In particular, the L2L optimization is carried out on a Linux-based host computer, whereas inner loop is simulated in its entirety on the later discussed neuromorphic hardware, Section~\ref{MMNMHW}.

\paragraph{Cross-entropy (CE) \citep{Rubinstein1997}} In each iteration, this algorithm fits a parameterized distribution $p(~\cdot~;\boldmath\phi)$ to the set of $n$ best-performing hyperparameters in terms of maximum likelihood. In the subsequent step, new hyperparameters are sampled from this distribution and evaluated. Afterwards, the procedure starts over again until a stopping criterion is met. Through this process, the algorithm tries to find a region of individuals where the performance is high on average. We used a univariate Gaussian distribution with a dense covariance matrix.

\paragraph{Evolution strategies (ES) \citep{Rechenberg1973}} In each iteration, this algorithm maintains base hyperparameters $\mathbf{\Theta}$ which are perturbed by random deviations $\boldmath\epsilon$ to form a new set of $n$ hyperparameters. This set is then evaluated and ranked by their fitness. In a subsequent step, the perturbations are weighted according to their rank to produce a direction of increasing fitness, which is used to update the base hyperparameters. Similar to Cross-entropy, ES also finds a region of hyperparameters with high fitness, rather than just a single one. Note that many variations of this algorithm have been proposed that differ for example in the way how the ranking or how the perturbations are computed \citep{Salimans2017}. In particular, we used Algorithm 1 from \citep{Salimans2017}.

\paragraph{Simulated annealing (SA) \citep{Kirkpatrick1983}} In each iteration, the algorithm maintains hyperparameters $\Theta$ and a temperature $T$. The hyperparameters are perturbated with a random $\epsilon$, whose size depends on the temperature $T$, and are evaluated later. The fitness of the unperturbed hyperparameters $\Theta$ is then compared with the perturbated hyperparameters $\Theta'$. The $\Theta'$ replaces $\Theta$ with a probability of $\min(1, \exp(\frac{-(f_{\Theta'} - f_{\Theta})}{T}))$. In the next step, the temperature is decreased following a predefined schedule and the new hyperparameters get perturbed. In contrast to the other methods discussed before, a single set of hyperparameters is the result. In our experiments, we simultaneously perform a number of parallel SA optimizations, using a linear temperature decay.

\paragraph{Numerical gradient-descent (GD)} In each iteration, the algorithm maintains hyperparameters $\Theta$ which are perturbed randomly in many directions and then evaluated. Subsequently, the gradient is numerically estimated and an ascending step on the fitness landscape is performed.

\subsection{Reinforcement Learning Problems}
\label{MMRL}
\begin{figure}[H]
\centering
\fbox{\includegraphics[width=180mm]{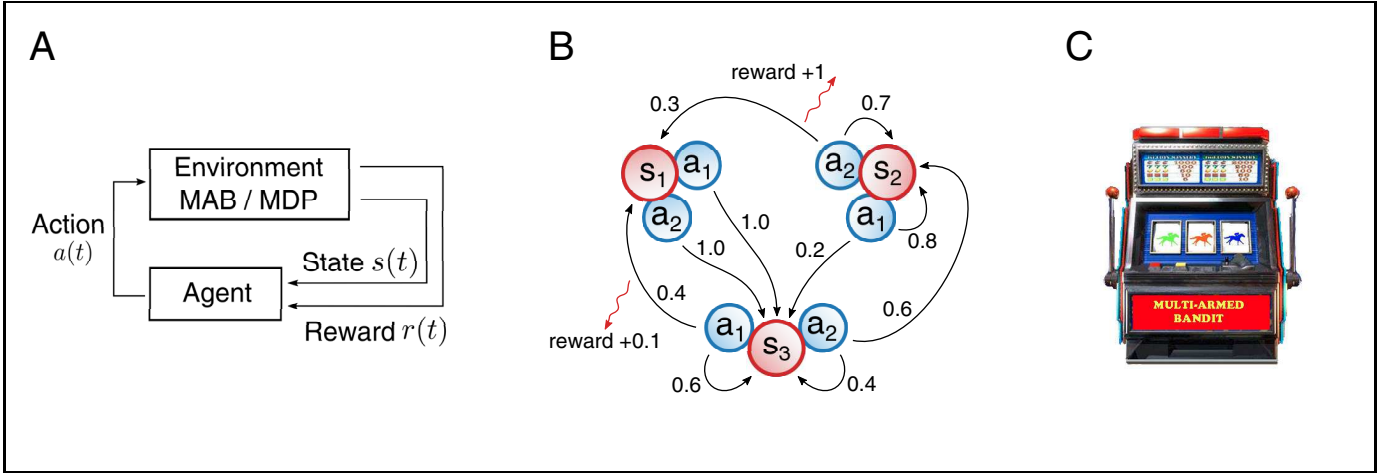}}
\caption{\textbf{Structure of reinforcement learning problems with examples for Markov Decision Processes (MDPs) and Multi-armed bandits (MABs)}. \textbf{(A)} General structure for reinforcement learning problems. An agent interacts with an environment in a loop. The agent selects an action based on state observations and receives a reward. \textbf{(B)} Example MDP with three states and two actions. State transitions are marked as arrows with annotated transition probabilities. A reward for a particular transition is indicated by a red arrow with the reward value along. Transitions with a probability of $0$ or rewards with a value of $0$ are omitted. \textbf{(C)} Illustration of a MAB: Bandit arms can be pulled that produce a reward stochastically.}
\label{fig:MMMDP}
\end{figure}

In all our experiments we considered reinforcement learning problems. Tasks of this type usually require many trials and sophisticated algorithms in order to produce a well-performing agent, since a teacher signal is only available in the form of a scalar quantity, the reward. To the worse, a reward does not arrive at every time step, but is often given very sparsely and only for certain events. Figure \ref{fig:MMMDP} (A) depicts a generic reinforcement learning loop. The agent observes the current state $s(t)$ of the environment and has to decide on an action $a(t)$. In particular, the agent samples an action according to policy $\pi(a | s)$, which is a probability distribution over actions $a$ given a state $s$. Upon executing the action, the environment will advance to a new state $s(t+1)$ and the agent receives a reward $r(t)$. In all our experiments, the RL environment was simulated on the neuromorphic chip.

\subsubsection{Markov Decision Process}
\label{MMMDP}
Markov Decision Processes (MDPs) are a well-known and established model for decision making processes in literature. A MDP is defined by a five-tuple $(\mathbb{S}, \mathbb{A}, p, r, \gamma)$, with $\mathbb{S}$ representing the state space, $\mathbb{A}$ the action space, $p$ the state transition function, $r$ the reward function and $\gamma$ a discount factor that weights future rewards differently from present ones. In particular, we are concerned here with such MDPs that exhibit discrete and finite state and action spaces. In addition, rewards are given in the range of $[0, 1]$. Figure \ref{fig:MMMDP} (B) shows a simple example of such a MDP with $||\mathbb{A}|| = 2$ and $||\mathbb{S}|| = 3$.\\

\noindent The goal of solving a MDP is to find a policy actions that yields the largest discounted cumulative reward $R$ that is defined as:
\begin{align}
R = \sum_{t=0}^{T} \gamma^t r(t) \label{eq:ResDiscountedCumReward}
\end{align}

\noindent In order to perform well on MDPs, the agent has to keep track of the rewarding transitions and must therefore represent the transition probabilities. Furthermore, the agent has to make a trade-off between exploring new transitions and consolidating already known transitions. Such problems have been studied intensively in literature and a mathematical framework was developed to optimally solve them \citep{Bellman1954}. The so-called Value-Iteration (VI) algorithm emerged from this framework and yields an optimal policy. Therefore, this algorithm is considered as the optimal baseline in all following MDP results.\\

\noindent In order to apply the L2L scheme, we introduce a family of tasks consisting of MDPs with a fixed size of the action and the state space. MDPs of that family are generated according to the following sampling procedure: whenever a new task is required, the rewards $r$ and the transition probabilities $p$ are randomly sampled from the range $[0, 1]$. In addition, the elements of $p$ are normalized such that the outgoing probabilities for all actions in each state sum up to 1.\\

\noindent We report our results in the form of a normalized discounted cumulative reward, where we scale between the performance of a random action selection and the performance of an optimal action selection, given by a policy produced by VI.

\subsubsection{Multi-armed Bandits}
\label{subsec:mab}
As a second category of RL problems, we consider multi-armed bandit (MAB) problems. A MAB is best described as a collection of several one-armed bandits, each of which produces a reward stochastically when pulled. In other words, one can view MAB problems as MDPs with a single state and multiple actions. Despite the deceptive simplicity of such problems, a great deal of effort was made in science to study these problems and the celebrated result of~\citep{gittins} showed that a learning strategy exists.\\

\noindent For the sake of brevity, we use the same notations for MABs as for MDPs. In particular, we say that the environment is always in one state $s_1$ and the agent is given the opportunity to pull several bandit arms $i$, which corresponds to actions $a_i$. In all experiments regarding MABs, we considered two-armed bandits, where each bandit produces a reward of either $0$ or $1$ with a fixed reward probability $p_i$. We investigate the impact of L2L on the basis of two different families of MAB tasks:
\begin{enumerate}
    \item \textit{unstructured bandits}: A task of this family is generated by sampling each of the two reward probabilities $p_1, p_2$ independently and uniform in $[0, 1]$.
    \item \textit{structured bandits}: A task of this family is generated by sampling the reward probability $p_1$ uniformly in $[0, 1]$ and compute $p_2 = 1 - p_1$.
\end{enumerate}

\noindent Similar to MDPs, we report our results for MABs in the form of a normalized cumulative reward, where we scale between the performance of a random action selection and the performance of an oracle that always picks the be best possible bandit arm. As a comparison baseline, we employ the Gittins index policy and note that the computation of the Gittins index value is calculated in the same way for both families. In particular, the Gittins index values are calculated assuming that the reward probabilities are independent (unstructured bandits).

\subsection{Neuromorphic hardware - HICANN DLSv2}
\label{MMNMHW}
\begin{figure}[h]
    \centering
    \fbox{\includegraphics[width=180mm]{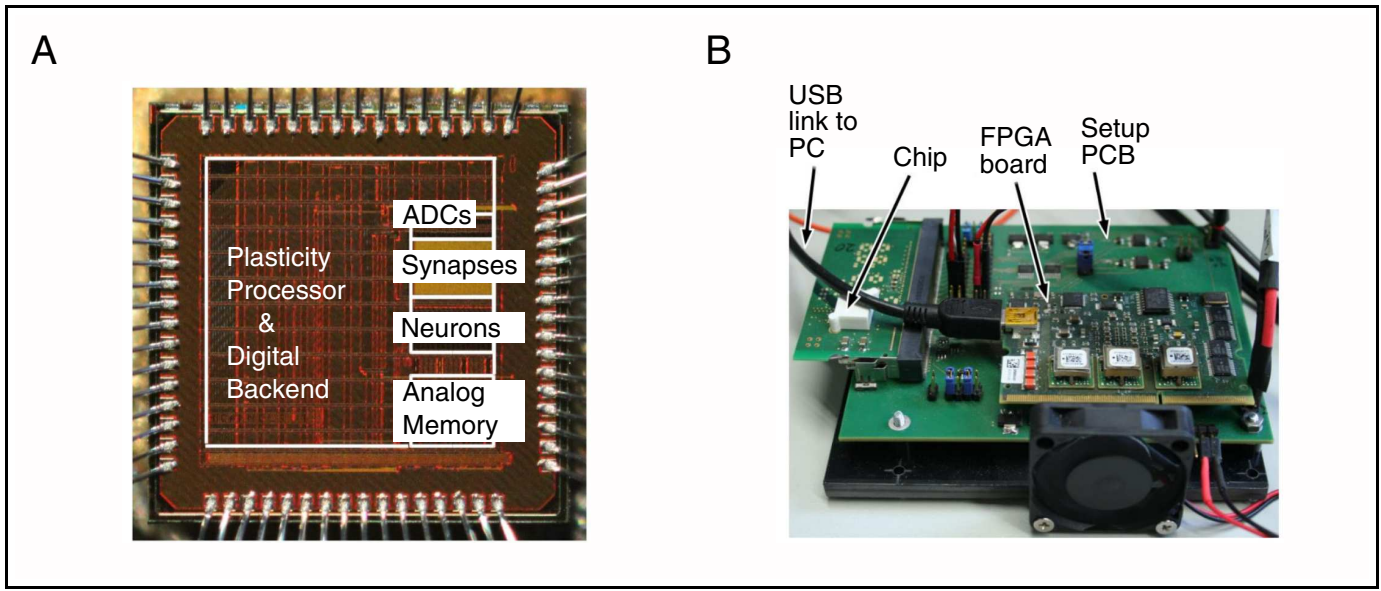}}
    \caption{\textbf{Neuromorphic chip micrograph and measurement setup adopted from \citep{Aamir2016a}.} \textbf{(A)} Micrograph of the neuromorphic hardware. The plasticity processing unit, the area responsible for the synaptic part, the neuronal part, a memory area as well as analog to digital converters (ADCs) are marked. \textbf{(B)} Measurement setup and prototype board. The board shows the neuromorphic chip itself, the interface to the host computer and a supportive FPGA board.}
    \label{fig:aamir2016}
\end{figure}

\noindent Various approaches for specialized hardware systems implementing spiking neural networks emerged and fundamentally differ in their realizations, ranging from pure digital over pure analog solutions using optical fibers up to mixed-signal devices \citep{Indiveri2011, Nawrocki2016, Schuman2017}. Whereas every single such NM hardware comes with certain advantages and limitations, one promising platform is the HICANN-DLS \citep{Friedmann2017}, herein used in the prototype version 2.\\

\noindent The hardware is a prototype of the second generation BrainScaleS-2 system currently under development as part of the Human Brain Project neuromorphic platform \cite[]{hbp_markram}. It represents a scaled-down version of the future full-size chip and is used to evaluate and demonstrate new features as illustrated in this work.\\

\noindent Conceptually the chip is a mixed-signal design with analog circuits for neurons and synapses, spike-based, continuous time communication and an embedded microprocessor. The NM hardware is realized in a 65nm CMOS process node by the company TSMC. It features 32 neurons of the leaky-integrate-and-fire (LIF) type connected by a 32x32 crossbar array of synapses such that each neuron can receive inputs from a column of 32 synapses. Synaptic weights can be set with a precision of 6-bits and can be configured row-wise to deliver excitatory or inhibitory inputs. Synapses feature local short-term (STP) and long-term (STDP) plasticity. All analog time constants are scaled down by a factor of 1000 to represent an accelerated neuromorphic system compared to biological time-scales, a feature that is strongly exploited in this paper.\\

\noindent The embedded microprocessor is a 32-bit CPU implementing the Power-PC instruction set with custom vector extensions. It is primarily used as a plasticity processing unit (PPU) for arbitrary operations on synaptic weights and labels. As a general purpose processor, it can also act on any other on-chip data like neuron and synapse parameters as well as on the network connectivity. It can also send and receive off-chip signals like rewards or other control signals. We make use of the freely-programmable PPU and investigated different learning algorithms which are explained in Section \ref{MMLA}. They all exploit the proposed network structure from Section \ref{MMNWStructure} and have the commonality, that the reward information of the state transitions is encoded in the synaptic efficacy.\\
In addition to learning algorithms, the plasticity processing unit also allows implementing environments for an agent. Since the system features a high speedup factor, any environment must also provide the same speedup factor, but with this closed-loop setup, the full potential of the neuromorphic hardware is unlocked.\\

\noindent Some of the basic design rationales behind the second generation BrainScaleS-2 system with special emphasis on the PPU are described in \citep{Friedmann2017}. Figure \ref{fig:aamir2016} (A) shows the measurement setup and (B) shows the micrograph of the hardware. In addition to other components, the measurement setup hosts the neuromorphic chip, a USB-Interface to connect the baseboard with a host computer as well as a separate FPGA board to control the experiments. The micrograph of the neuromorphic chip shows the different components and where they are located. A description of the actual prototype used in this work including details on the neuron implementation and the synaptic array can be found in \citep{Aamir2016a}.

\subsection{Network Structure and Action Selection}
\label{MMNWStructure}
\begin{figure}[ht!]
    \centering
    \fbox{\includegraphics[width=180mm]{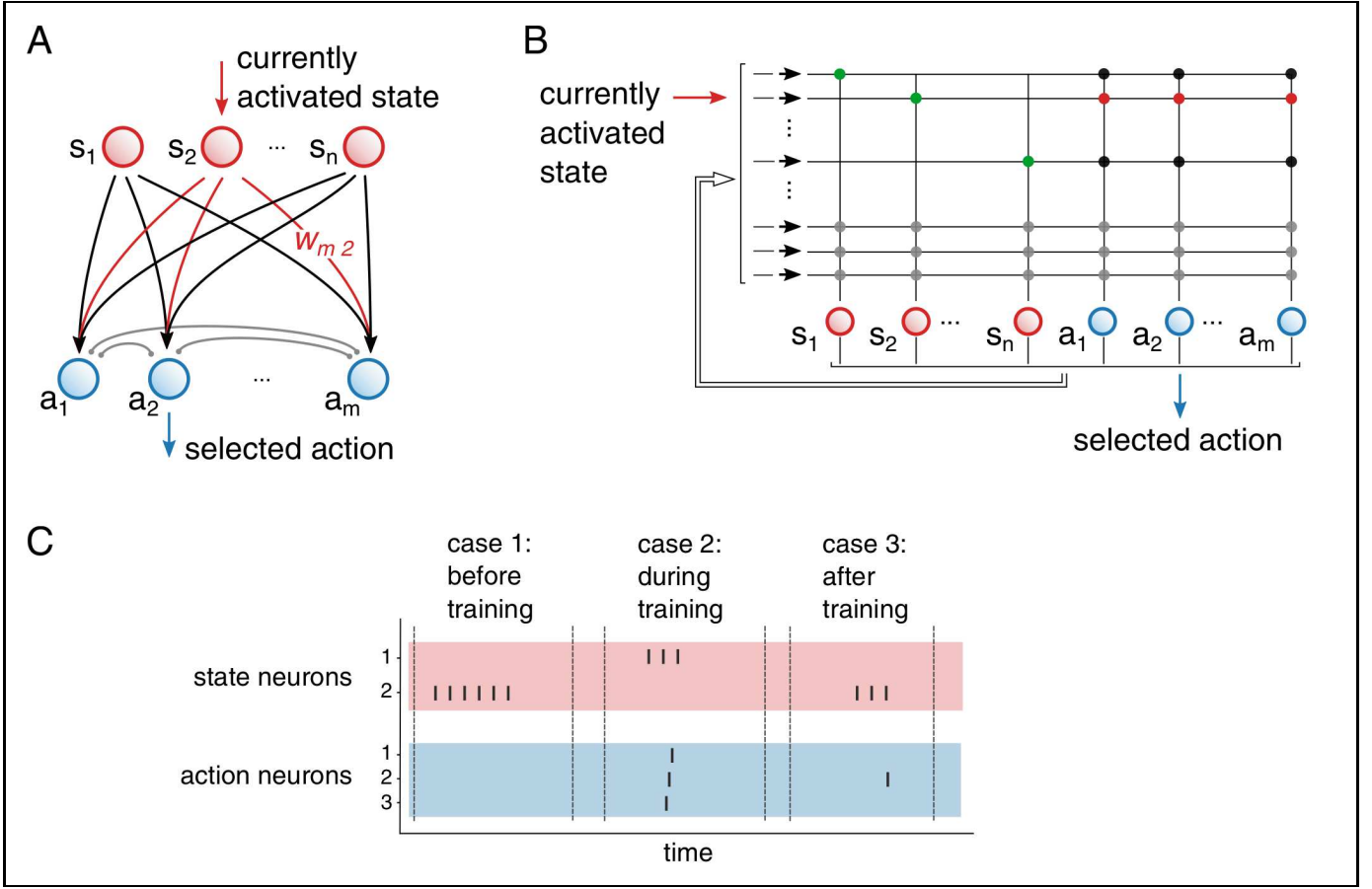}}
    \caption{\textbf{Neural network structure and realization on neuromorphic hardware}. \textbf{(A)} Network structure with two populations: state population (red), action population (blue). Excitatory synapses $w_{ij}$ (black and red) are plastic and used for learning. Inhibitory synapses (gray) introduce mutual inhibition in the action population. \textbf{(B)} Mapping of the network onto the neuromorphic hardware. Synapses are organized in crossbar array of size (32x32). We use autapses (green) for persistent exication of state neurons. Persistent excitation is stopped by additional inhibitory synapses that connect the action population to the state population.
    \textbf{(C)} Three examples of the action selection process. In case 1, none of the action neurons received enough input to emit a spike: A random action is selected. In case 2, each action neuron emits a spike: A random action among active neurons is selected. In case 3, only a single neuron of the action population emits a spike that determines the selected action.}
    \label{fig:methodsNWStructure}
\end{figure}

\noindent Like discussed in Section~\ref{MMRL}, the agent is required to select an appropriate action $a(t)$ given a particular state $s(t)$ of the environment. We discuss in this Section how the agent can be implemented using a network of spiking neurons on neuromorphic hardware. Since our experiments were concerned with either Multi-armed bandits or Markov Decision Processes, we designed the network structure for the more general MDP problems. In particular, the design is based on the Markov Property of MDPs, using the fact that the next state $s(t+1)$ solely depends on the chosen action $a(t)$ and the current state $s(t)$, similarly to \citep{Friedrich2016a}.\\

\noindent Concretely, we make use of a feed-forward network of spiking neurons with two populations, as illustrated in Figure \ref{fig:methodsNWStructure} (A). One population encodes the state of the environment (state population, marked in red) and the second population encodes all possible action choices (action population, marked in blue). We assume that all states exhibit the same number of possible actions. Under this assumption, the resulting agent commits to specific actions by the following action selection protocol: Given that the agent finds itself in state $s_j$, then the corresponding state neuron receives stimulating input and produces output spikes that are transmitted to the neurons $a_i$ of the action population by excitatory synapses $w_{ij}$. Eventually, this stimulation will trigger a spike in the action population, depending on the synaptic strengths $w_{ij}$. The action $a(t)$ that will be taken, is determined by the neuron of the action population that emits a spike first. In addition, neurons coding for actions are connected inhibitory among each other with synapses of strength $\xi$, through which a WTA-like network structure arises. Due to this mutual inhibition, mostly a single neuron of the action population will emit a spike and hence, trigger the corresponding action.\\

\noindent In practice, additional tricks are required to implement the proposed scheme on the neuromorphic device, see Figure~\ref{fig:methodsNWStructure} (B). To continually excite the active state neuron, we send a single spike that triggers a persistent firing through strong excitatory autapses (marked in green). If a neuron from the action population eventually emits a spike, the active state neuron needs to be prevented from further spiking. For this purpose, we use inhibitory synapses of strength $\zeta$ projecting from action neurons to state neurons. Due to synaptic delays, more than one action neuron may emit a spike. In such a case, an action is randomly selected among the set of active neurons. It is to be noted that smaller inhibition weights lead to more random exploration, because insufficient inhibition will not prevent spikes of other action neurons, in which case action selection becomes randomized.\\

\noindent One other implementation detail comes from the fact that there is only a limited resolution (6 bit) available on the NM hardware. This might cause that weights saturate at either $0$ or the maximum weight value and prevent efficient learning. To avoid this problem, the weights $w_{ij}$ are rescaled with a certain frequency $f_\mathrm{rescale}$ according to:

\begin{align}
    k &= \frac{W_\mathrm{max} - W_\mathrm{min}}{\max(w_{ij}) - \min(w_{ij})}\\
    d &= W_\mathrm{max} - k \max(w_{ij})\\
    w_{ij}' &= k w_{ij} + d
\end{align}

\noindent where $W_\mathrm{max}$ and $W_\mathrm{min}$ provide the upper and lower rescale boundary.\\

\noindent Figure~\ref{fig:methodsNWStructure} (C) depicts typical examples of the action selection process for three common cases occurring throughout the learning process. In case 1 (usually before training), a state neuron, i.e. corresponding to state 2, is active and persistently emits a spike. However, none of the synapses connecting to the action neurons is strong enough to cause a spike. In such a case, after a predefined time, the state neuron is inhibited and a random action is selected. In case 2 (likely during learning), another state neuron is active, but all synapses to the action neurons are strong enough to cause every action neuron to spike before the mutual inhibition sets in. In such a case, a random action among the active action neurons is selected. Eventually, the system reaches case 3 (after learning), where only a single action neuron is excited by a given state neuron.\\

\noindent Learning in this network structure is implemented by synaptic plasticity rules that act upon the excitatory weights $w_{ij}$ projecting from the state to the action population. In particular, these weights pin down which action has the highest priority for each state.\\
\subsection{Learning Algorithms}
\label{MMLA}
\subsubsection{Q-Learning}
\label{subsubsec:QLearning}
MDPs have been studied intensively in computer science and a rigorous framework on how to solve problems of this kind optimally was introduced by Bellman in \citep{Bellman1954}.
An important quantity in MDPs is the so-called Q-Function, or Action-Value function. The Q-Function $Q^\pi(s, a)$ expresses the expected discounted cumulative reward, when the agent starts in state $s$, takes action $a$ and subsequently proceeds according to its policy $\pi$. Formally, one writes this as:

\begin{align}
Q^{\pi}(s_j, a_i) &= \mathds{E}\left[\sum^{\infty}_{k=0}\gamma^k r(t+k+1) | s(t) = s_j, a(t) = a_i\right]
\end{align}
\noindent where $\gamma$ is the discount factor of the MDP and $r(t)$ is the immediate reward at time step $t$.
As discussed before in Section \ref{MMMDP}, we consider only discrete MDPs and the Q-Functions can therefore be represented in a tabular form. 
This property suits our network structure, since the synapses that project from the state population to the action population $w_{ij}$ can represent all Q-values $Q^\pi(s_j, a_i)$. Hence, we define $w_{ij} \stackrel{def}{=} Q^\pi(s_j, a_i)$.

\noindent To solve MDPs, the goal is to determine the optimal policy $\pi^*$. A common approach is to infer the Q-Function of an optimal policy $Q^*$ and then reconstruct the policy according to:
\begin{align}
    \pi^*(a | s) &= \begin{cases} 1 & \text{if}~a = \argmax_{a'} Q^*(s, a') \\ 0 & \text{else}\end{cases}
\end{align}

\noindent Indeed, as we aim to encode Q-values in synaptic weights $w_{ij}$, we emphasize that the $\argmax$ operation will be naturally carried out by the spiking neural network, as proposed in Section~\ref{MMNWStructure}. To infer the Q-values of the optimal policy, we derive rules of synaptic plasticity based on temporal difference algorithms as proposed by~\citep{Sutton1998}.

\paragraph{TD(1)-Learning}
\noindent Temporal Difference Learning (TD(1)-Learning) was developed as a method to obtain the optimal policy. The estimate of the optimal Q-Function is improved based on single interactions with the environment. TD(1)-Learning is guaranteed to converge to the correct solution as shown in \citep{Watkins1992, Dayan1994}. Based on TD(1), the synaptic weight updates take on the following form:

\begin{align}
w_{ij}(t + 1) &= w_{ij}(t) + \alpha \left( r(t) + \gamma \max_k w_{kj}(t) - w_{ij}(t)  \right) \quad \text{for}~s(t)=s_j, a(t)=a_i \label{eq:QTD}
\end{align}

\noindent Where $\alpha$ denotes a learning rate.
\paragraph{TD($\lambda$)-Learning}
\noindent The convergence speed of TD(1)-Learning can be further improved if one uses additional eligibility traces $e_{ij}(t)$ per synapse. The resulting algorithm is then referred to as TD($\lambda$)-Learning. In particular, the trace $e_{ij}$ indicates to what extent a current reward makes the earlier visited state-action pair $(s_j, a_i)$ more valuable. Convergence proofs of TD($\lambda$) are given in \citep{Dayan1992, Dayan1994}. To implement the algorithm, we update eligibility traces at every time step $t$ according to the schedule

\begin{align}
e_{ij}(t) &=
\begin{cases} 
\hfill \gamma \lambda e_{ij}(t-1) + 1 \hfill & \text{ if $s(t) = s_j$ and $a(t) = a_i$} \\
\hfill \gamma \lambda e_{ij}(t-1) \hfill & \text{ otherwise} \\
\end{cases}
\end{align}

\noindent In addition, we define an error $\delta(t)$ according to
\begin{align}
\delta(t) &= r(t) + \gamma \max_k w_{kj}(t) - w_{ij}(t) \quad \text{for}~s(t)=s_j, a(t)=a_i
\end{align}

\noindent which enables us to express the resulting plasticity rule as a product of the eligibility trace and error $\delta$. This update is carried out for every synapse $w_{ij}$:
\begin{align}
w_{ij}(t+1) &= w_{ij}(t) + \alpha \delta(t) e_{ij}(t) \quad \text{for all}~i, j\label{eq:QTDLambda}
\end{align}

\noindent The parameter $\lambda \in [0, 1]$ controls how many state transitions are taken into account and one obtains as one corner case TD(1)-Learning.

\subsubsection{Meta-Plasticity}
\label{MMMetaPlasticity}

\begin{figure}[t!]
    \centering
    \fbox{\includegraphics[width=85mm]{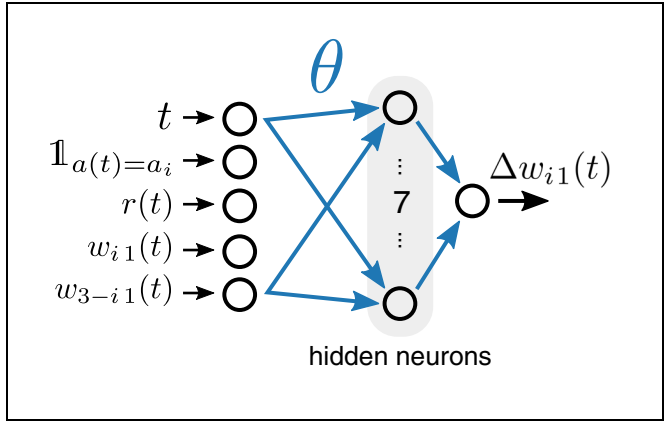}}
    \caption{\textbf{Meta-plasticity for a two-armed bandit task}. The plasticity rule is represented by a parametrized multi-layer perceptron with one hidden layer (denoted as ANN). It receives as inputs the time step $t$, a binary flag $\mathds{1}_{a(t)=a_i}$ that indicates if the weight to be updated was responsible for the selected action, the obtained reward $r(t)$, as well as the weights $w_{i\,1}$ and $w_{3-i\,1}$.}
    \label{fig:meta_methods}
\end{figure}

In order to tailor the specific update rule towards the actual task family at hand, we approached the problem also from the perspective of meta-plasticity. I.e. we represent the synaptic weight update by a parameterized function approximator. We then optimize its parameters with L2L in such a way that a useful learning rule for a given task family emerges. We used a multilayer perceptron, the architecture of which is visualized in Figure~\ref{fig:meta_methods}. The perceptron receives five inputs, computes seven hidden units with sigmoidal activation and provides one output, the weight update $\Delta w_{ij}$. Effectively, the input to output mapping of this approximator is specified by a number of free parameters $\theta$ (weights of the multilayer perceptron) that are considered as hyperparameters and optimized as part of the L2L procedure. Since the multilayer perceptron is a type of an artificial neural network, this plasticity rule is referred to as ANN learning rule. The update of synaptic weights $w_{ij}$ thus takes on the general form of:

\begin{align}
w_{ij}(t+1) &= w_{ij}(t) + f_\mathrm{ANN}(\mathrm{inputs}_{ij}(t); \theta)
\end{align}

\noindent The specific choice of inputs is salient for the possible set of learning rules that can emerge. In the case of the ANN learning rule, we only considered structured MAB, where each of the two synapses is updated at every time step. We set the inputs in this case to a vector
\begin{align}
    \mathrm{inputs}_{i\,1}(t) &= \begin{pmatrix} t\\ \mathds{1}_{a(t)=a_i}\\ r(t)\\ w_{i\,1}(t)\\ w_{3-i\,1}(t) \end{pmatrix}
\end{align}
that is composed of the current time step $t$, a binary flag indicating if the synapse was responsible for the action at step $t$, the obtained reward $r(t)$, the weight $w_{i\,1}(t)$ and the weight of the synapse associated to the other bandit arm $w_{3-i\,1}(t)$.

\subsection{Analysis of Meta-Plasticity}
\label{MMAnalMetaPlasticity}
Since we use an artificial neural network in our meta-plasticity approach, which can represent fairly general nonlinear functions, it is hard to understand what the arising plasticity rule actually expresses. To investigate the emerged functionality, one approach used in literature is called functional Analysis of Variance (fANOVA) which was presented in \citep{Hutter2014}. This method originally aims to assess the importance of hyperparameters in the machine learning domain. It does so by fitting a random forest to the performance data of the machine learning model that was gathered using different hyperparameters.\\
\\
\noindent We adopted this method but applied it to a slightly different, but related problem. Our goal is to assess the impact of each input of the ANN rule with respect to its output. To do so, the weights $\Theta$ of the plasticity network remain fixed, while the input values to the plasticity network as well as the output from the plasticity network are considered as inputs to the fANOVA framework. Based on this data, a random forest with 30 trees is fitted and the fraction of the explained variance of the output with respect to each input variable can be obtained. 

\section{Results}
This section presents the results of our approach implemented on the described neuromorphic hardware. First, we report how L2L can improve the performance and learning speed in Section \ref{RLTL}. Then, we investigate the impact of outer loop optimization algorithms in Section \ref{RLTLAL} and demonstrate in Section \ref{RMetaPlas} that Meta-Plasticity yields competitive performance, while also enhancing transfer learning capabilities. Finally, we investigate the speedup gained from the neuromorphic hardware by comparing our implementation on the NM hardware to a pure software implementation of the same model in Section \ref{RCompT}.

\subsection{Learning-to-Learn improves Learning Speed and Performance}
\label{RLTL}
Here, we first demonstrate the generality of our network structure when applied to Markov Decision Processes. Then, we examine the effects of an imposed task structure more closely by investigating Multi-armed Bandit problems. To efficiently train the network of spiking neurons, we employed Q-Learning and derived corresponding plasticity rules, as described in see Section \ref{subsubsec:QLearning}. The plasticity rules itself, as well as the concrete implementation on NM hardware, are influenced by hyperparameters $\Theta$ that we optimized by L2L, such that the cumulative discounted reward for a given family of tasks is improved on average, see Section \ref{MMLTL}.\\

\noindent We implemented a neuromorphic agent that learns MDPs. In fact, the proposed network structure in Section \ref{MMNWStructure} is particularly designed for such tasks and we applied concretely TD($\lambda$), see equation~\eqref{eq:QTDLambda}.
Hyperparameters included all occurring parameters of the employed TD($\lambda$)-Learning rule $\alpha, \gamma, \lambda$, the inhibition strength among the action neurons $\xi$, the strength of inhibitory weights connecting the action neurons to the state neurons $\zeta$, as well as the variables influencing the hardware-specific rescaling $f_\mathrm{rescale}, W_\mathrm{max}$ and $W_\mathrm{min}$. Therefore, the complete hyperparameter vector was given as $\Theta = (\alpha, \gamma, \lambda, \xi, \zeta, f_\mathrm{rescale}, W_\mathrm{max}, W_\mathrm{min})$. We used the discounted cumulative reward, Equation~\ref{eq:ResDiscountedCumReward}, as the fitness function $f(C; \Theta)$ and optimized $\Theta$ using CE.
We used a batch size of $N=20$.\\

\begin{figure}[ht!]
\centering
\fbox{\includegraphics[width=.5\textwidth]{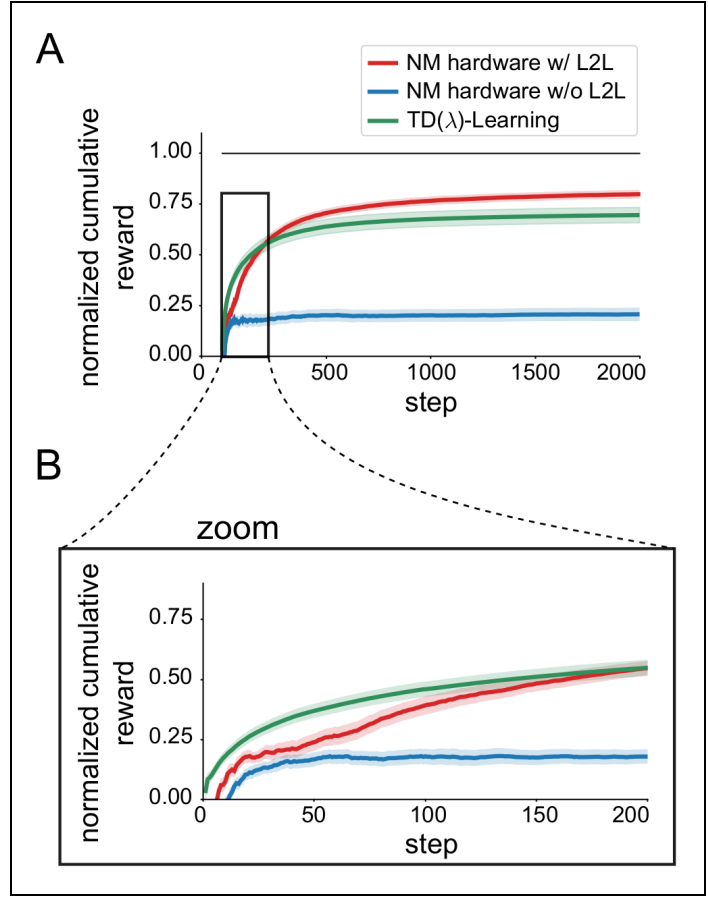}}
\caption{\textbf{Impact of L2L for Markov Decision Processes}. \textbf{(A)} Average learning performance on the MDP task family ($||\mathbb{S}|| = 2, ||\mathbb{A}|| = 4$) using TD($\lambda$)-Learning, see Equation~\eqref{eq:QTDLambda}. Learning performance is expressed as the normalized cumulative discounted reward ($0$ random, $1$ optimal) and is averaged over $50$ different tasks. Shaded areas mark the uncertainty of the mean. \textbf{(B)} Zoom into the first $200$ steps to emphasize increased learning speed.}
\label{fig:ResQLam}
\end{figure}

\noindent The results for the MDP tasks are depicted in Figure \ref{fig:ResQLam} (A) where we report the discounted cumulative reward for $T = 2000$ steps. The discounted cumulative reward is normalized in such a way, that VI is scaled to $1$ and the random policy is scaled to $0$. To compare with, we used a TD($\lambda$)-Learning implementation from a software library\footnote{https://pymdptoolbox.readthedocs.io/en/latest/index.html} without a spiking neural network (green line).\\
We found that applying L2L improved the discounted cumulative reward (red solid line), compared to the case where the hyperparameters are randomly chosen (blue line). In addition, the learning speed was also increased, which can be seen in the zoom depicted in Figure \ref{fig:ResQLam} (B).\\

\FloatBarrier

\begin{figure}[ht!]
    \centering
    \fbox{\includegraphics[width=85mm]{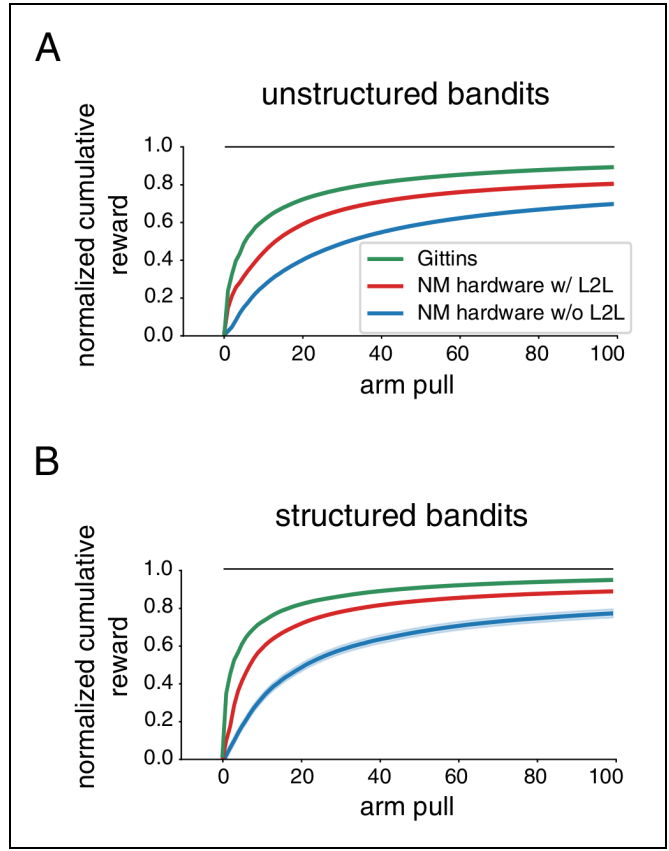}}
    \caption{\textbf{Impact of L2L for Multi-armed Bandit tasks}.
    We report the normalized cumulative reward ($0$ random, $1$ optimal) on MAB tasks averaged over $1000$ different tasks. The shaded areas mark the uncertainty of the mean. The neuromorphic agent uses TD(1)-Learning, see Equation~\eqref{eq:QTD}. \textbf{(A)} \textit{unstructured bandits} \textbf{(B)} \textit{structured bandits}
    }
    \label{fig:ltl_mab}
\end{figure}

\noindent In the case of MABs, we focused on small networks and two arms in the bandit, which allowed us to complement the results that were obtained for general MDPs of larger size. We considered two families of MABs: \textit{unstructured bandits} and \textit{structured bandits} (\ref{subsec:mab}) which the neuromorphic agent had to learn using the TD(1)-Learning rule, see Equation~\eqref{eq:QTD}, where we set $\gamma = 1$. In addition, we introduced here a decaying learning rate defined as $\alpha(t) = \alpha_\mathrm{decay} \alpha_0$, using a decay factor of $\alpha_\mathrm{decay}$. We then used L2L to carry out a hyperparameter optimization separately for both MAB families and optimized the parameters of the TD(1)-Learning rule $\alpha_0$ and $\alpha_\mathrm{decay}$, the inhibition strength among action neurons $\xi$ and the inhibitory weights of synapses that connect the action population to the state population $\zeta$. Hence, the hyperparameter vector was given as $\Theta = (\alpha_0, \alpha_\mathrm{decay}, \xi, \zeta)$. We used the cumulative reward as the fitness function $f(C; \Theta)$ and optimized $\Theta$ using CE. We used a batch size of $N=200$.\\

\noindent In Figure~\ref{fig:ltl_mab} we report the performance results that were obtained before and after applying L2L. The agent interacted for $T = 100$ steps with a single MAB and we compare with a baseline given by the Gittins index policy, as described in \ref{subsec:mab}. We found that after performing a L2L optimization the performance was enhanced, which was even more apparent for \textit{structured bandits}. In particular, L2L endowed the agent with a better learning speed, which is exhibited by a faster rising of the performance curve. This can only be achieved when the hyperparameters of the learning system are well tailored to the tasks that are likely to be encountered, which was the responsibility of L2L.
We also observed that the agent could still learn a MAB task to a reasonable level even if no L2L optimization was carried out. This is implied by the fact that TD(1)-Learning is primed to learn RL tasks. However, this also raises the question of how well such a general plasticity rule can adapt to the level of variations exhibited by analog circuitry. We consider extensions in Section~\ref{sec:meta}.

\subsection{Performance comparison of Gradient-free Optimization Algorithms in the outer loop}
\label{RLTLAL}
\begin{figure}[H]
\centering
\fbox{\includegraphics[width=85mm]{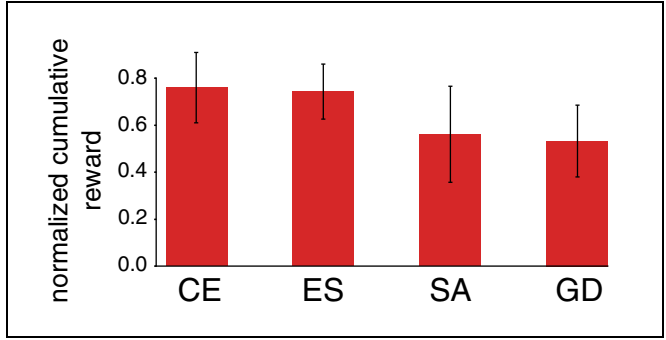}}
\caption{\textbf{Performance impact of different outer loop algorithms.} We exhibit the performance of an L2L optimized neuromorphic agent on MDP tasks with $||\mathbb{S}|| = 2$ and $||\mathbb{A}|| = 4$. The performance is measured as the final normalized discounted cumulative reward after $T = 2000$ steps and are averaged over $50$ different tasks. We compare Cross-Entropy (CE), Evolution strategies (ES), Simulated annealing (SA) and numerical Gradient descent (GD), as described in Section~\ref{MMLTL}. The dimensionality of the hyperparameter vector was 8, as in Section~\ref{MMMDP}.}
\label{fig:ResOuterloopImpact}
\end{figure}

The results presented so far suggest that the concept of L2L can improve the overall performance and also lays the foundation that abstract knowledge about the task family at hand is integrated into an agent. However, the choice of a proper outer loop optimization algorithm is also crucial for this scheme to work well. The modular structure of the L2L approach used in this paper allows to interchange different types of optimization algorithms in the outer loop for the same inner loop task. To demonstrate the impact in terms of performance when using different optimization algorithms, several such algorithms were investigated for both general MDPs and also for specialized MAB tasks. Figure~\ref{fig:ResOuterloopImpact} shows a comparison of the final discounted cumulative reward at the end of the tasks for different outer loop optimization algorithms.

\noindent Depending on the inner loop task considered, we found that the cross-entropy (CE) method, as well as evolution strategies (ES), work well because both aim to find a region in the hyperparameter space, where the fitness is high. This property is particularly desired when it comes to noise in the fitness landscape due to imperfections of an underlying neuromorphic hardware. In addition, both can cope with noisy fitness evaluations and do not overestimate a single fitness evaluation which could easily lead to a wrong direction in the presence of high noise in the fitness landscape.\\

\noindent However, a simpler algorithm such as simulated annealing (SA) can also find a hyperparameter set with rather high fitness. Especially when running multiple separate annealing processes in parallel with different starting points, the results can almost compete with the ones found by CE or ES. However, SA does not aim at finding a good parameter region but just tries to find a single good set of working hyperparameters. This is prone to cause problems because a single good set of hyperparameters offers less robustness compared to an entire region of well-performing hyperparameters. A simple numerical gradient-based approach did not yield good results at all because of the noisy fitness landscape. In general, the developer is free to choose any optimization algorithm in the outer loop when using L2L. New algorithms can also be implemented which are specially tailored to a particular problem class, which can lead to a new research direction.

\subsection{Performance improvement through Meta-Plasticity}
\label{RMetaPlas}
\label{sec:meta}
\begin{figure}[t!]
    \centering
    \fbox{\includegraphics[width=180mm]{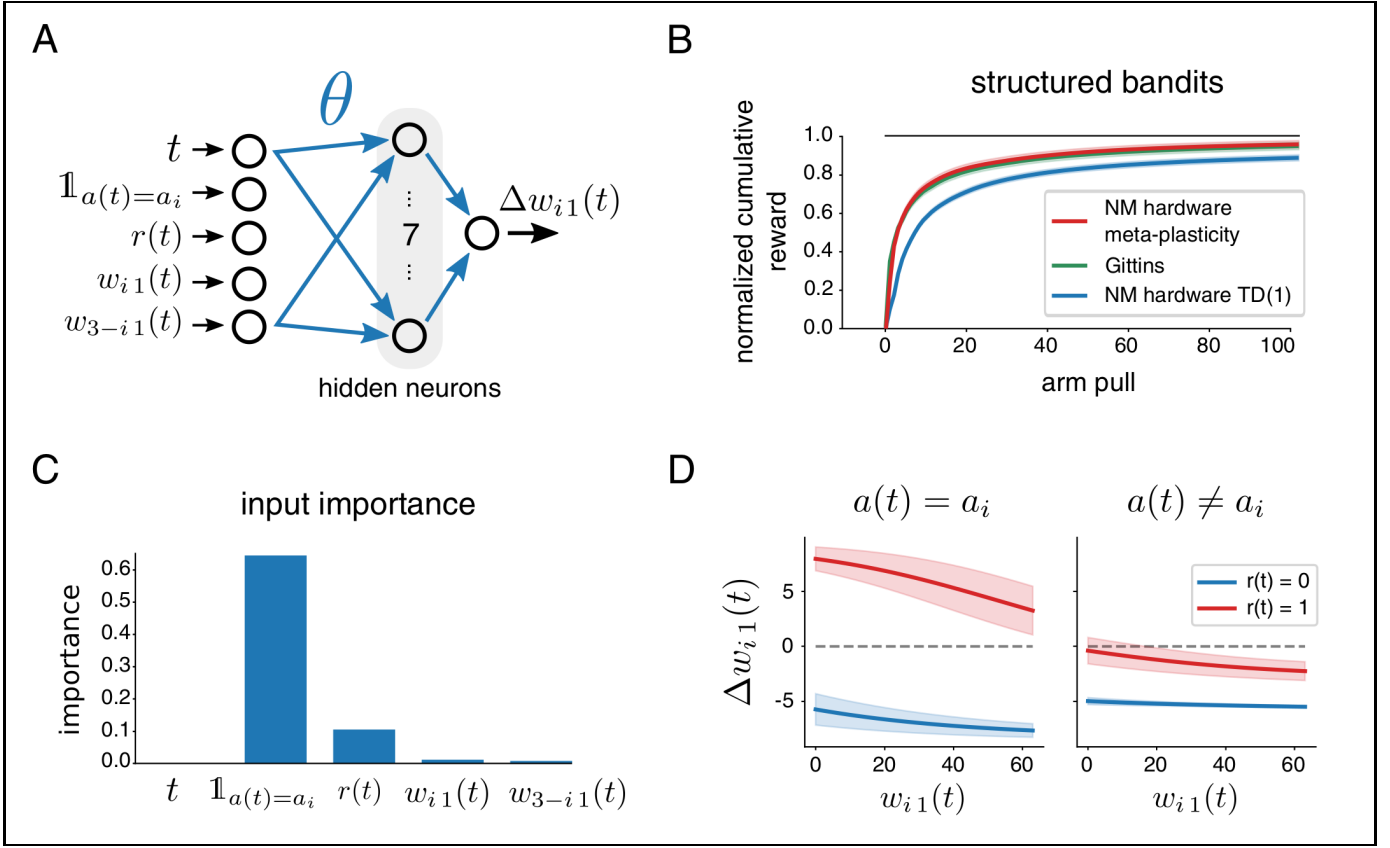}}
    \caption{\textbf{Meta-plasticity for a two-armed bandit task}. \textbf{(A)} The plasticity rule is represented by a parametrized multi-layer perceptron. \textbf{(B)} Performance of the meta-plasticity as compared to an optimized TD(1)-Learning neuromorphic agent and Gittins index on structured MABs. \textbf{(C)} Relative contributions of each input to the variance of the weight update as computed by fANOVA. Mostly responsible are the action flag and the reward signal. \textbf{(D)} The weight update as shown for the different possible cases of $\mathds{1}_{a(t) = a_i}$ and $r(t)$ depending on the current weight $w_{i\,1}$. Shaded areas indicate effects of inputs that are not fixed by the variables on the axes: $t$ and $w_{3-i\,1}$.
    }
    \label{fig:metaplasticity}
\end{figure}
We further asked the question if one can install plasticity rules in spiking neural networks that allow to improve learning from rewards beyond the level of Q-Learning on specific tasks. We conjectured that the TD(1)-Learning rule that was used on MAB tasks offers a somewhat limited surface for incorporating task structure or imperfections in analog hardware. Hence, we pursued a different approach to implement learning from rewards using meta-plasticity. I.e. we specified the entire learning rule hyperparameters. In particular, we used a multilayer perceptron of 7 hidden neurons as envisioned in Figure~\ref{fig:metaplasticity} (A), whose input to output behavior implements the plasticity rule of synapses in the spiking neural network. This is apparently the first example of meta-plasticity on neuromorphic hardware, where a rule for synaptic plasticity is evolved through optimization by L2L.\\

\noindent To test the approach, we used L2L to optimize all occurring hyperparameters on the task family of \textit{structured bandits}. In particular, the hyperparameter vector was composed of the parameters of the plasticity rule $\theta$ and the inhibition strengths $\xi$ and $\zeta$: $\Theta = (\theta, \xi, \zeta)$. We used the cumulative reward, Equation~\ref{eq:ResDiscountedCumReward}, as the fitness function $f(C; \Theta)$ and optimized $\Theta$ using CE with a batch size of $N=200$. \\

\noindent We summarize our results in Figure~\ref{fig:metaplasticity} and observed a drastic increase in learning performance of agents employing the evolved learning rule~(B). As opposed to the results for a tuned version of the TD(1)-Learning rule, we can achieve a performance that is on the same level as the Gittins index policy. This highlights that the tailor-made plasticity rule for a family of tasks can counteract negative effects of imperfections in the neuromorphic hardware.\\
\\
\noindent Even though the arising learning rule performs well on average on the family of tasks it has been trained on, there is no theoretical guarantee for that. Hence, an analysis of the optimized learning rule was conducted, where we examined the importance of the multiple inputs provided to the update rule for the resulting output, see Figure~\ref{fig:metaplasticity}~(C). Apparently, the most important inputs are the flag that represents if the current weight was responsible for the last action and the obtained reward. Since both of the inputs can assume only two values, one can visualize the four different cases in four different curves. We report the expected weight change depending on the current weight, see Figure~\ref{fig:metaplasticity}~(D), where we average over other unspecified inputs.\\

\noindent Updates for weights which were responsible for the previous action are in the direction of the obtained rewards. Hence, the meta-plasticity rule reinforced actions depending on the reward outcome, similarly to Q-learning rules. Interestingly however, the update of the synaptic weight which had not caused the last action was always negative independently of the reward. We believe that L2L simply found that it does not matter what happens to the weight that did not cause actions, because as long as it does not increase, it will not disturb the current belief of the best bandit arm.\\
\begin{figure}
    \centering
    \fbox{\includegraphics[width=180mm]{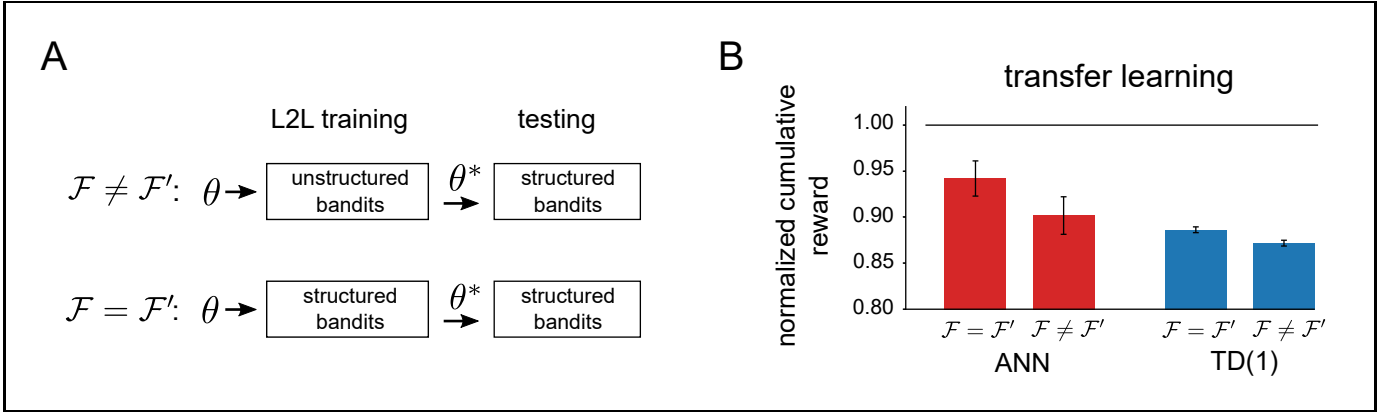}}
    \caption{\textbf{Transfer-learning}. \textbf{(A)} Setup to investigate transfer learning capabilities. \textbf{(B)} Comparison of transfer-learning capabilities of different learning rules. The meta-plasticity update rule (ANN) can potentially encode more task structure through a greater dimensionality of the hyperparameter vector $\Theta$.}
    \label{fig:transfer_learning}
\end{figure}
\\
\noindent To test if the reinforcement learning agent on the neuromorphic hardware has been optimized for a particular range of tasks, we carried out another experiment. We tried to answer if the agent can take advantage of the abstract task structure if it was present. To do so, we always tested learning performance on \textit{structured bandits}, denoted as $\mathcal{F}'$. For optimization with L2L, we instead used either \textit{unstructured bandits} or \textit{structured bandits}, and we denote the family on which hyperparameter optimization was carried out by $\mathcal{F}$. This experimental protocol (Figure~\ref{fig:transfer_learning}~(A)) allowed us to determine to which extent abstract task structure can be encoded in hyperparameters. We report the results for neuromorphic agents in Figure~\ref{fig:transfer_learning}~(B), where we considered the TD(1)-Learning rule and the meta-plasticity learning rule. Consistently, we observed that optimizing hyperparameters for the appropriate task family enhances performance. However, we conjecture that the greater adjustability of the meta-plasticity learning rule renders it to be better suited for transfer learning as compared to TD(1)-Learning rule.

\subsection{Exploiting the benefit of accelerated hardware for L2L}
\label{RCompT}
One of the main features of neuromorphic hardware devices is the ability to simulate spiking neural networks very fast and efficient. To make this more explicit for the MDP tasks, a software implementation with the same network structure and the same plasticity rule was conducted on a standard desktop PC using one single core of an Intel\textsuperscript{\texttrademark} Xeon\textsuperscript{\texttrademark} CPU X5690 running at 3.47 GHz. The spiking neural network was implemented using the Neural Simulation Technology (NEST) \citep{Gewaltig:NEST} with a Python interface and the plasticity rule as well as the environment were also implemented in Python. 
To have a better comparison, two families of MDP tasks with different sizes of $||\mathbb{S}||$ and $||\mathbb{A}||$ were defined. The first family is defined by $||\mathbb{S}|| = 2$ and $||\mathbb{A}|| = 4$ (small MDP) and the second family by $||\mathbb{S}|| = 6$ and $||\mathbb{A}|| = 8$ (large MDP).\\

\begin{figure}[H]
\centering
\fbox{\includegraphics[width=180mm]{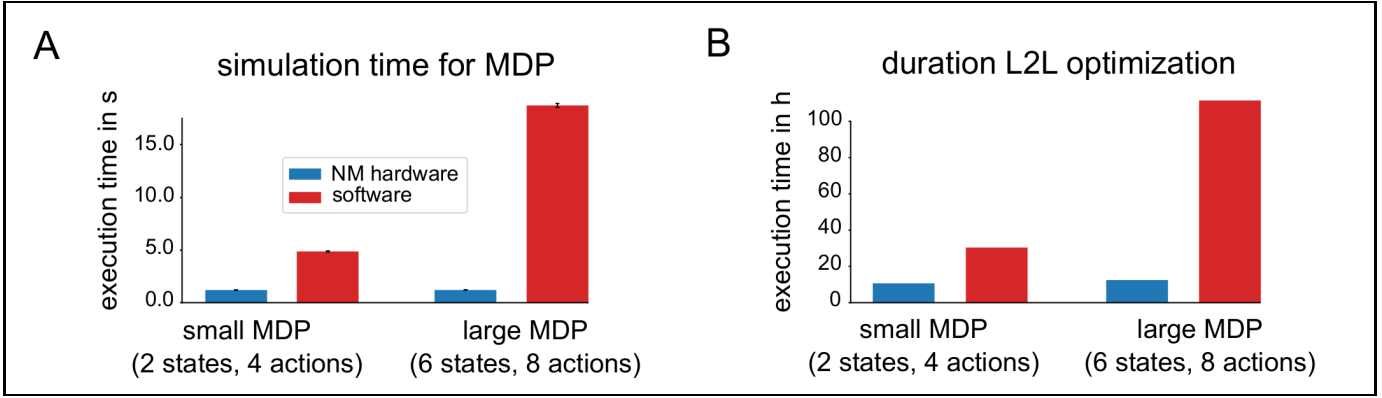}}
\caption{\textbf{Impact of accelerated neuromorphic hardware on simulation time}. \textbf{(A)} Shows the comparison of the required simulation time, averaged over $50$ different MDP tasks of two different sizes. In software simulations, only a single CPU core was used. The simulation time of the NM hardware is shorter and remains constant for the two families. \textbf{(B)} Shows the duration comparison for an entire L2L optimization for two different families.}
\label{fig:ResNMHWSpeedup}
\end{figure}

\noindent Figure \ref{fig:ResNMHWSpeedup} (A) shows a comparison of the simulation time needed for a single randomly selected MDP tasks, averaged over $50$ MDPs and for each of the two families. The simulation times include implementation specific overheads, for example, the communication overhead with the neuromorphic hardware.
One can see that the simulation time needed for MDP tasks with both sizes are shorter using the neuromorphic hardware and in addition, the simulation time needed to solve the larger task does not increase. First, this indicates, that the neuromorphic hardware can carry out the simulation of the spiking neural network faster and second, that using a larger network structure does not yield an additional cost, as long as the network can fit on the NM hardware. In contrast to this, using more neurons requires longer simulation times in pure software. 
A similar key message can be found in Figure \ref{fig:ResNMHWSpeedup} (B), where instead of a single MDP run, an entire L2L run is evaluated on the neuromorphic hardware as well as with the software implementation. Both, the L2L run on neuromorphic hardware as well as the one in software can in principle be easily parallelized when using more hardware systems or more CPU cores which would decrease the overall simulation time. Note that scheduler overheads are not taken into considerations.\\

\section{Discussion}
\label{sec:discussion}
Outstanding successes have been achieved in the field of deep learning, ranging from scientific theories and demonstrators to real-world applications. Despite impressive results, deep neural networks are not out of the box suitable for low-power or resource-limited applications. Instead, spiking neural networks are inspired by the brain, an arguably very power efficient computing machine. The neuromorphic hardware that we used in this work was designed to port key aspects of the astounding properties of this biological circuitry to silicon devices.\\

\noindent The human brain has been prepared by a long evolutionary process over with a set of hyperparameters and learning algorithms that can be used to cover a large variety of computing and learning tasks. Indeed, humans are able to generalize task concepts and port them to new, similar tasks, which provides them with a tremendous advantage as compared to most of the contemporary neural networks. In order to mimic this behavior, we employed gradient-free optimization techniques, such as the cross-entropy method or evolutionary strategies (see Section \ref{MMLTL}), applied in a Learning-to-Learn setting. This two-looped scheme combines task-specific learning with a slower evolutionary-like process that results in a good set of hyperparameters as demonstrated in Section \ref{RLTL}. The approach is generic in the sense that both, the algorithms mimicking the slower evolutionary process and the learning agent can be exchanged. In principle, any agent with learning capabilities can be used as the learning agent and any optimization algorithms as the evolutionary process. We found that some outer loop optimization algorithm perform better than others and the optimization algorithms should ideally be chosen with the inner loop task in mind. Outer loop optimization algorithms need to operate in a high-dimensional parameter space, have the ability to deal with noisy result evaluations, have the ability to find a good final solution and also require a low number of parameter evaluations before reaching a good solution. Algorithms that aim to find a region of hyperparameters with high performance such as evolution strategies or cross-entropy worked the best for us, see Section \ref{RLTLAL}.\\

\noindent L2L offers both, either to find optimal hyperparameters for a fixed individual task or to boost transfer learning capabilities of an agent when using a family of tasks. In addition, new optimization algorithms can be developed to further improve performance in the outer loop of L2L. In this work, we used reinforcement learning problems in connection with NM hardware to demonstrate the aforementioned benefits.\\

\noindent In particular, the concept of L2L allows to shape highly adjustable plasticity rules for specific task families. The usage is not only limited to spiking neural networks but can also be applied to artificial neural networks. This may yield potential for a future research direction. Apparently, this is the first time that the idea of L2L and Meta-Plasticity was applied to a neuromorphic hardware, see Section \ref{RMetaPlas}.\\

\noindent Neuromorphic hardware allows to emulate a spiking neural network with a significant speedup compared to the biological equivalent, which makes a large number of computations, required in the L2L scheme, feasible. To quantify the overall speedup of the accelerated neuromorphic hardware, a comparison with a pure software simulation on a conventional computer was carried out (see Figure \ref{fig:ResNMHWSpeedup}). We conclude that the two-looped L2L scheme is especially suited for accelerated neuromorphic hardware.\\

\section*{Conflict of interest statement}
The authors declare that the research was conducted in the absence of any commercial or financial
relationships that could be construed as a potential conflict of interest.

\section*{Author Contribution}
WM, TB and FS developed the theory and experiments. TB implemented and conducted experiments with regard to MDPs, benchmarked performance impact of outer loop optimization algorithms and probed the performance benefit of NM hardware. FS implemented and conducted experiments with regard to MABs. FS, CP and WM conceived meta-plasticity, FS and CP implemented it. FS tested the benefits in transfer learning. TB, FS, CP, WM and KM wrote the paper. 

\section*{Funding}
This research/project was supported by the HBP Joint Platform, funded from the European Union’s Horizon 2020 Framework Programme for Research and Innovation under the Specific Grant  Agreement  No.   785907  (Human  Brain  Project  SGA2).

\section*{Acknowledgments}
We thank Anand Subramoney for his support and the contributions to the Learning-to-Learn framework. We are also grateful for the support during the experiments with the neuromorphic hardware. In particular, we like to thank David St{\"o}ckel, Benjamin Cramer, Aaron Leibfried, Timo Wunderlich, Yannik Stradmann, Christian Mauch and Eric M{\"u}ller. Furthermore, we also like to thank Elias Hajek for useful comments on earlier versions of the manuscript.

\bibliographystyle{apalike}
\bibliography{references}

\end{document}